\documentclass[journal,transmag]{IEEEtran}

\usepackage{color,graphicx}
\usepackage{float}

\begin{document}
 
\title{Cooperative Automated Vehicles: a Review of Opportunities and Challenges in Socially Intelligent Vehicles Beyond Networking}

\author{\IEEEauthorblockN{Seng W. Loke
\IEEEauthorblockA{School of Information Technology, Deakin University, Geelong, Victoria, Australia}
\thanks{Manuscript received XXX XX, 2018; revised XXX XX, 2018. 
Corresponding author: S.W. Loke (email: seng.loke@deakin.edu.au).}
}}


\maketitle

\begin{abstract}
The connected automated vehicle has been often touted as  a technology that will become pervasive in society in the near future. 
One can view an automated vehicle as having Artificial Intelligence (AI) capabilities, being able to self-drive, sense its surroundings, recognise objects in its vicinity, and perform reasoning and decision-making.
 Rather than being stand alone, we examine the need for automated vehicles to cooperate and interact within their socio-cyber-physical   environments, including the problems cooperation  will solve, but also the issues and challenges. We review current work in cooperation for automated vehicles, based on  selected examples from the literature. {\color{black} We conclude noting the need for the ability to behave cooperatively as a form of  {\em social-AI} capability for automated vehicles, beyond sensing the immediate environment and beyond the underlying networking technology.}  
\end{abstract}

\section{Introduction}


Connected automated vehicles (CAVs), forming the so-called Internet of Vehicles~\cite{Yang2017,Datta2017VehiclesAC}, are predicted to revolutionise transportation world-wide, and transform urban life, as early as 2021, and becoming pervasive in the decades to come.\footnote{See https://www.bbhub.io/dotorg/sites/2/2017/05/\-TamingtheautomatedVehicleSpreadsPDF.pdf,\\http://www.businessinsider.com/companies-making-driverless-cars-by-2020-2016-11/?r=AU\&IR=T/\#tesla-made-a-big-move-this-year-to-meet-its-goal-of-having-a-fully-self-driving-car-ready-by-2018-1, around the world testing of automated vehicles (see http://insuranceblog.accenture.com/where-in-the-world-are-self-driving-cars), NHTSA guidelines on development of automated vehicles: https://www.nhtsa.gov/technology-innovation/automated-vehicles\#voluntary-guidelines, and self-driving vehicle initiative in Australia (http://advi.org.au/)} Other forms of vehicles such as self-driving wheelchairs\footnote{https://spectrum.ieee.org/the-human-os/biomedical/devices/selfdriving-wheelchairs-debut-in-hospitals-and-airports}, and self-driving motorcycles\footnote{E.g., see https://www.yamaha-motor.com.au/discover/design-lab/Motobot and https://www.dezeen.com/2017/01/12/honda-unveils-self-balancing-motorcycle-design-transport-ces/} are also being developed. {\color{black} Cooperative Intelligent Transport Systems (e.g.,  cooperative driving) is an active area of research.\footnote{For example, see  https://www.car-2-car.org/about-c-its/ on the Car 2 Car communication consortium, https://tca.gov.au/car/c-its, and https://www.toyota-global.com/innovation/intelligent\_transport\_systems/infrastructure/}}

A lot of work has been on sensors for vehicles and how the Artificial Intelligence (AI) in vehicles can learn to ``€œsee''€, navigate and manoeuvre within everyday road systems, {\color{black} e.g., see~\cite{machines5010006}}.
In the larger  Internet-of-Things (IoT) service ecosystems and the emerging cognitive IoT~\cite{6766209},  CAVs are situated within  socio-technical environments of human road users and other automated entities.
 In so far as the action of CAVs must take into account the actions and reactions of others, and are intentional individually or collectively, CAVs need to perform social interactions and social signalling.

Vehicles need to interact with and potentially connect not just to other vehicles, but also motorcycles, bicycles, pedestrians, and other road-users, as well as with  IoT services (including via Road-Side-Units), over Dedicated Short Range Networking (DSRC) or 5G-V2X networking~\cite{10.1007/978-3-319-72329-7_1}. 
Indeed, there have been much research on vehicle-to-vehicle (v2v) and vehicle-to-infrastructure (v2i) (and more generally, v2x) communications. Over such network protocols, there is opportunity for vehicles to exchange application level messages and cooperate to improve safety and  increase their effectiveness~\cite{7932586},
creating a {\em cooperation layer} above the vehicular {\em network layer}. 
The Society of Automotive Engineers (SAE) released a message set dictionary for standardizing messages exchanged in DSRC communications, such as intersection collision warnings, emergency vehicle alerts and vehicle status information
 can be shared.\footnote{At https://www.sae.org/standardsdev/dsrc/ and in particular the message set dictionary SAE J2735 at https://saemobilus.sae.org/content/j2735\_200911}  There are large European projects on cooperative-ITS including cooperative vehicles.\footnote{E.g., http://c-mobile-project.eu/}
 The European Telecommunication Standard Institute
(ETSI) provided the EN 302 637-2 standard\footnote{http://www.etsi.org/deliver/etsi\_en/302600\_302699/30263702/\-01.03.01\_30/en\_30263702v010301v.pdf} which defined  Cooperative Awareness Messages (CAMs).

 CAVs will not just get from A to B autonomously but, in doing so, will need to {\em cooperate} with other vehicles and people in a wide range of situations.  The notion that vehicles have social capabilities in that they can negotiate, cooperate and collaborate has been proposed by Riener and Ferscha~\cite{riener}.
 Here, 
 we  explore further the idea of social vehicles, envisioning the   {\em social brain of a CAV}, which is defined as   a software module that determines how a vehicle  cooperates with other vehicles, how a vehicle  cooperates with pedestrians and services over such v2x networking, how a vehicle  reasons about social behaviours, how a  vehicle  behaves when receiving particular messages,  and how inter-vehicular cooperation can be exploited in road situations. In particular, the social brain of a CAV could  reason about social situations with respect to other vehicles on the road and pedestrians, remember interactions in the past to inform future cooperation, work within social norms for the road, and have context-aware focus (akin to salience in humans in social cognition). {\color{black} However, the social brain is not merely reasoning about situation-awareness but also determines the actions of the vehicle - how the vehicle will behave socially with respect to other vehicles and how the vehicle interacts (e.g., what messages to send) with other entities.}

We use the metaphor of the {\em social brain} in humans~\cite{doi:10.1098/rstb.2009.0160}, a functional aspect of the human brain which relates to understanding others, empathising, trusting, communicating and cooperating with others. Similar to the  social brain in humans, we can consider the notion of a social brain in a CAV which allows it  to predict or model what other vehicles and people are going to do,  understand the intentions of other vehicles and people, and behave in a way that exhibits that  understanding in a wide range of situations. 

Computationally, the social brain in a vehicle could implement multiple cooperation protocols for different
 road situations, e.g., a protocol for collision warning at intersections, a protocol for movement at roundabouts, a protocol for merging traffic, a protocol for platooning on the highway, a protocol for overtaking, a protocol for cars giving way at intersections, and so on, all integrated  into a  {\em social brain} module in the vehicle.  At the same time, policy rules and robot laws are needed to govern not just road traffic in general but to ensure the  trustworthy and ethical interaction between CAVs, and between CAVs and people~\cite{doi:10.1177/0739456X16675674}.

This paper aims to outline the {\em cooperation layer} in CAVs, highlighting potential applications and issues of cooperative CAVs, drawing on related work.
We first describe a range of scenarios where cooperation can happen as well as the potential benefits, and then outline challenges  in enabling the social brain of CAVs.

{\color{black}
The contribution of this paper is twofold: (1) to review work on cooperative vehicles highlighting the wide range of cooperative behaviours being investigated for vehicles, and (2) to propose the notion of the social brain for autonomous vehicles, i.e., broadly speaking, a ``{\em social-AI}'' for vehicles, which functions as the central component in supporting a wide range of cooperative behaviours.   
}

\section{Types of Cooperation}

We consider below two types of cooperation: vehicle-to-vehicle,  and
vehicle-to-pedestrian.

\subsection{Vehicle-to-Vehicle Cooperation and Reasoning}
We  consider a range of  scenarios, including parking, routing,  managing traffic flows, cooperative awareness, and social memory,
 to illustrate the potential that v2v cooperation in addressing important on-road problems.
 Vehicle-to-vehicle cooperation will use v2v networking but could also be mediated by v2x infrastructure.

{\bf Parking.}
Cooperation is not merely sharing of information. For example,
the use of inter-vehicle cooperation in finding parking spaces have been investigated in different aspects.  The authors in~\cite{Kokolaki:2012:OAP:2161001.2161219} analysed the performance of cooperative vehicles in finding parking spaces through collecting and sharing  parking information compared with centralized management under different conditions and concluded  that there is no optimal solution for all situations and the benefit of sharing information can outweigh the increased  vehicle competition that it may cause. Also, the authors in~\cite{7457417} found through simulations that disseminating parking information among vehicles barely decreased the searching time and even occasionally increased the searching time. The work in~\cite{Bessghaier2012} aimed to reduce contention among vehicles by restricting sharing to only information that the vehicle itself is not interested in.   In~\cite{7580732} is discussed a decentralised car parking allocation mechanism inside a car park by supporting the vehicles with initial information about available slots at the car park gate and using vehicular cooperation that shares   intentions about where to park and negotiates to resolve competition - a  reduction of time-to-park of up to 25\% is possible. Hence, it is not necessarily just sharing information, but cooperation via negotiation and coordination. There are other notions of interaction among cars when parking (e.g., a car leaves a note for another saying ``let me know when you leave'').

{\bf Routing.}
Recent work in vehicle-to-vehicle cooperation 
 explored vehicles coordinating their routes, and in doing so, can 
 distribute themselves along faster, even if longer, routes. For example,
 {\color{black}
the work in~\cite{5730496} uses an ant behaviour inspired approach for  an environment-centric coordination mechanism for large numbers of vehicles, where  information is shared analogous to ant pheromones, achieving indirect coordination via stigmergy. 
 }
 
In~\cite{DBLP:journals/tits/DesaiLDS13}, by having cars opportunistically cooperating on routes, over DSRC v2v communication, 
 when they come near each other at disparate intersections, 
traffic congestion across a vast area can be alleviated; vehicles can get to destinations sooner by as much as 30\% compared to all just taking the shortest distance route. It is noteworthy that some gains can be obtained  without global cooperation among all vehicles, but merely by local cooperation within groups of  vehicles   at intersections. This suggests some cooperation even locally where possible can have far reaching consequences. The work in~\cite{10.1371/journal.pone.0159110} showed that inter-vehicle communication can help to detect congestion and to suggest new routes to avoid congestion.

{\color{black} 

 In~\cite{4434989}, a decentralised approach for routing of multiple Unmanned Aerial Vehicles (UAVs) is presented, where a range of sites appear dynamically and need to be visited (or served) by a UAV; each UAV computes its destination using an algorithm based on a solution to the Dynamic Traveling Repairperson Problem (DTRP), resulting in a spatial distribution of work.

Eco-routing to reduce fuel consumption, emissions, and travel time has been explored from different perspectives, as an optimization problem~\cite{8317643}, and using a recommendation engine based on historical information about fuel usage over road segments, and real-time traffic information~\cite{6236175}.  Very interestingly, v2i communications can be used for exchanging eco-routing information (e.g., fuel consumption costs of a route) enabling vehicles to select the best route for
a trip~\cite{7313112}.

In~\cite{7479108}, the routing game, based on non-cooperative game theory, has been explored for multiple agents to learn about the behaviour of other agents in order  to make routing decisions about the use of limited shared resources (e.g., roads). Indeed, such analysis aims to provide models to predict flow distribution over routes. In a game theoretic model~\cite{7604118}, route allocation of certain controlled agents is determined, given the learning behaviour  of the other (selfish) agents. Such work is motivated by the idea that selfish agents, each seeking to maximise its own utility (e.g., choosing a route to minimize its own latency) results in suboptimal performance system-wide, and so, some control of part of the traffic flow can be useful. Such work are suggestive of the benefits of cooperative approaches. 

}

{\bf Swarm Behaviours for Dynamic Traffic Flows.}
Traditionally,   markings on the road or traffic signals are used to coordinate vehicles so that they move in an orderly manner, but they can cause delays or reduce the utility of roads. With cooperation and coordination, traffic flow can be coordinated without physical signals. 

\begin{figure*}
    \centering
    \includegraphics[width=13cm,height=6cm]{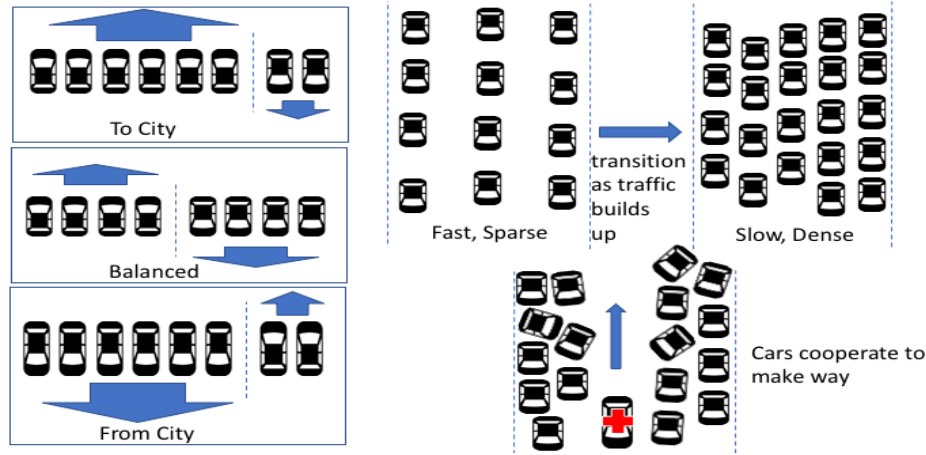}
    \caption{Swarm-Based Dynamic Traffic Flow}
    \label{swarm}
\end{figure*}

Cooperation among vehicles can also be used to form flexible collective vehicle behaviours. These possibilities are illustrated in Figure~\ref{swarm}. For example, on a highway,  lanes are fixed equally on both sides (e.g., five lanes in each direction) - while at certain times, traffic can be heavier in one direction than the other. 
With CAVs, when traffic volume is high in one direction,  vehicles could cooperate massively and inform other vehicles that there are eight lanes now in one direction and two in the reverse way. Lanes can then be re-balanced at other times.  
Also, some ``lanes'' can become narrower (with cars moving closer to each other but slower) at certain times, while at other times lanes are broader (with cars moving faster but further apart from each other).
This idea has been called ``traffic shaping'' in~\cite{riener}. 

Another scenario is making way for an emergency vehicle passing through - CAVs that receive notifications of the emergency vehicle can coordinate their actions to create a path way. All vehicles in the vicinity could receive the same message from the emergency vehicle, but each need to decide what is the best action to take so that a path is created, based on   contextual knowledge of itself and its surrounding vehicles.  The vehicles must also be able to detect that the emergency vehicle has passed through so that they can cooperatively resume their normal movements.

{\color{black}
Viewing the many vehicles as agents in swarms has also been proposed in~\cite{7553030}, where a vehicle swarm architecture is described. The swarm model for vehicles in~\cite{7328361} is based on inter-vehicle communications, i.e. Vehicular Ad Hoc Networks (VANETs), supporting interactions among the vehicles in swarms, where it is shown how obstacle-avoidance situations can be implemented as swarm behaviours, using a force model on each vehicle. Interestingly, the work in~\cite{10.1007/978-3-319-72823-0_58} uses the idea of fish swarms to model inter-vehicle spacing and clustering behaviours among vehicles, showing that with each vehicle sharing their status information (e.g., position, velocity, etc), vehicle group moving behaviour based on fish behaviour is possible.
}

{\bf Platooning, Intersections and Safety.}
Vehicles can opportunistically platoon to improve travel times, road usage, and   safely do so~\cite{7547317}.  Vehicle platooning on highways has been widely  explored~\cite{DBLP:journals/comsur/JiaLWZS16}. Cooperative adaptive cruise control is already explored~\cite{7934129}, where movement models of vehicles are shared via v2v communications, rather than raw data about locations.

Cooperation between vehicles entering a highway and vehicles on the highway can enable safer traffic merging, and can be achieved via vehicle intention recognition based on speeds~\cite{6629471} and v2v communication~\cite{8023664}. The idea of cars cooperating to facilitate traffic merging is also noted in~\cite{riener}.

{\color{black} The survey in~\cite{7562449} reviewed schemes for the coordination of connected automated vehicles for merging and at intersections,  via centralized schemes where there is a central controller and via decentralized schemes, based on v2v or v2i communications.

Indeed, there have been much work in enabling vehicles to pass through intersections efficiently and safely. Inter-vehicle communication have been shown to be useful to guide vehicles at intersections without traffic lights~\cite{4012536}. The work in~\cite{4357693} showed how vehicles approaching an intersection can send messages to other vehicles via an intersection agent in order to provide warnings to reduce collisions at  intersections. 
 Collision warning at intersections and at hard-to-see settings via v2v communication is  widely noted.\footnote{E.g., see https://www.nhtsa.gov/sites/nhtsa.dot.gov/files/readiness-of-v2v-technology-for-application-812014.pdf} Also, vehicle to bicycle and  bicycle to bicycle interactions can enhance safety~\cite{7929602}. The work in~\cite{5342530} showed that  v2v communications can be used by autonomous vehicles to determine the position
and speed of the surrounding vehicles at an intersection, and by using this to adjust their speeds based on fuzzy logic reasoning, the vehicles can reduce traffic jams 
 significantly and the number of cars going through the intersection can be maximized.
Wang~{\em et al.}~\cite{8317924} showed that cooperation among vehicles via
vehicle-to-vehicle communication as they move through together at a signalized intersection with reduced inter-vehicle gaps   can yield better outcomes - reducing unnecessary speed fluctuations and unnecessary stops.

The work in~\cite{6156445} explores a new   algorithm for an intelligent intersection, which can   receive and confirm reservation requests from vehicles approaching the intersection, and recommend a speed profile for vehicles
to follow when passing through  the intersection. This provides a level of coordination among vehicles crossing such an intersection allowing reductions in delays and increased traffic handling.
The work in~\cite{7070682} studied a reservation-oriented  mechanism based on FIFO queues biased with priorities for scheduling automated vehicles with different priorities when passing through intersections. Again, such coordination achieved this way is useful for handling conflicting aims of the vehicles at intersections.

 MIT has been working on intersections without traffic lights  based on a slot-based system (based on Intersection Managers).\footnote{http://senseable.mit.edu/light-traffic/} 
 Also, in \cite{6907953}, a scheme is proposed where an Intersection Coordination Unit (ICU) is used to coordinate the smooth movement of automated vehicles through the intersection almost eliminating  stop delays. Two-way
communication is used to receive status information from  
approaching vehicles, such as current position, velocity, and destination, and the ICU then computes and sends guidance information to them.

Cooperative Adaptive Cruise Control (CACC) has been explored to allow vehicles to coordinate their movements relative to each other in order to pass through an intersection efficiently, with reduced delay and fuel consumption~\cite{6338827}. The work in~\cite{6824011} studied simple rules for each vehicle to behave as it arrives at an intersection, taking into account whether it has right-of-way and its distance to other vehicles, yielding efficient and safe intersection crossings. 

Linear programming can be applied to determine lane-based traffic flow models for autonomous vehicles passing through intersections without traffic lights, assuming such vehicles can communicate with very low latency with each other, perhaps mediated by intersection controllers~\cite{ZHU2015363}.  Similarly, the work in~\cite{6121907} proposed a method for autonomous vehicles to pass through intersections safely and efficiently, without traffic lights, assuming a central roadside infrastructure control system and perfect v2i  communications within a circle of 150m radius. This idea has been substantially extended in a 
simulation case study with four intersections~\cite{LEE2013193}. A coordinator is used in~\cite{DBLP:journals/automatica/MalikopoulosCZ18} to manage information sent among automated vehicles at a signal-free intersection, but the control decisions are made by the vehicles themselves in a distributed manner using the information, resulting in a scheme where energy consumption can be minimized (e.g., via reducing stops) while maximizing throughput at the intersection.

With autonomous vehicles, rather than just parking, a major aspect of their behaviour is to drop passengers off.
Autonomous vehicles could also form platoons and adjust their speeds relative to each other in order to drop passengers off efficiently, as shown in ~\cite{8370676}.

}

Interesting is the work in~\cite{DBLP:journals/corr/abs-1708-06374} which formalized rules for automated vehicle behaviour   to achieve safety (and comfort) goals. Such rules tend to govern vehicles in relation
 to how such vehicles interact safely with other vehicles and people - such safety rules relates not only to good social behaviour but could be encoded in regulations governing such vehicles.

An approach for vehicles to cooperate when they need to use the same road spaces is given in~\cite{DBLP:conf/itsc/ManzingerA18}. If two or more vehicles need to traverse the same road space, e.g., at roundabouts, intersections, or to avoid obstacles, the vehicles can negotiate via an auction to bid for the road space. Bids can be based on the utility of the road space to the vehicle, given its goal, reachable areas, and position. An interesting aspect about this approach is its generality,  usable in different road situations where conflicts on road areas occur, not just in a specific situation like intersections as in many approaches above.

{\bf Cooperative Awareness.} As vehicle share information with each other, on position, movements and status,
each vehicle can built up an integrated picture of other vehicles in a particular area, and increase situation-awareness can enable better decision-making. {\color{black} Indeed, cooperative awareness can help significantly improve each vehicle's awareness of the environment~\cite{machines5010006}.}
To enable such cooperative awareness, as mentioned, ETSI provided  Cooperative Awareness Messages (CAMs), the performance of which has been  studied, e.g.~\cite{10.1007/978-3-642-41054-3_3,8119570}.
The work in~\cite{10.4108/eai.31-8-2017.153052} investigated a mechanism where 802.11p v2v communications  was used for frequent updates.

CAM enables vehicles to know each other's position, dynamics and attributes, and react in response to this, but
the ability to aggregate and synthesise such information  to create a coherent model of the situation around a vehicle and for a vehicle to formulate the best action in the situation are challenges.
For example, suppose it is known to a vehicle-0 that vehicle-1 is moving in a certain direction and speed, vehicle-2 is moving in the opposite direction with a certain speed, and vehicle-3 is stationary along the path of vehicles 1 and 2, this could mean that a three vehicle collision is about to take place in front of vehicle-0, then how will vehicle-0 figure out that such a collision could take place within a certain time, and given its view of the road structure and knowledge of the type and structure of vehicles 1, 2 and 3, how could it compute its best action? In general,  aggregating and making sense of 
  information from disparate sources require sophisticated reasoning and the ability to make predictions about vehicle behaviour.

{\bf Long Term Cooperation - Social Networks and Social Memory.}
Future cars could have {\em social memory}.
Cars can record past pattern of interactions and connections with other cars and things, forming conceptual links with objects in the world, e.g., labelling some links as ``car-friend''€.  A car can remember the favours received by another car, and so, reciprocate when it gets a chance.

Such remembered links and relationships do not mean that cars cannot communicate with one another as needed or ad hoc (in the same that people have social networked friends but also can interact with anyone who are not in their social network, and add social network friends). The advantage of such remembered connections and relationships is to track regularity of pattern of interaction and perhaps even establish a trust network for information exchange, reciprocal (favour exchange) interactions, and to perform problem-solving over such networks. For example, cars   take turns to give way to one another or defer parking spots to one another at different times. The social memory provides a context for future v2v communications, yielding greater efficiency in expressing intent.  

The notion of the Social Internet  of Vehicles (SIoV) has been well elaborated in~\cite{Maglaras_2016,doi:10.1108/LHT-12-2017-0259}, towards new context-aware applications involving vehicles, drivers and passengers.  
{\color{black} The SIoV notion can mean a social network among people in vehicles, or among vehicles themselves, separate from social human networks.  For example, the work in \cite{luan2015feel} used v2v communication to overcome the limitations of Internet Connectivity for highway communications in order to to build a social network among passengers. In~\cite{6849025}, relationships   established between  vehicles and between the vehicles and the road side units (RSUs) are explored, including parental-object relationships between all vehicles of the same manufacturer and emerging in the same period, social-object relationships among vehicles that come into contact via v2v communications, and co-work-object relationships when vehicles meet and work with RSUs. The work in \cite{BUTT201868} proposed a hierarchical architecture for the SIoV, comprising a system of cloud, fog and edge nodes, and propose the use of Web standards for interoperability. As vehicles work and decide autonomously in the SIoV, there are ethical issues as reviewed in~\cite{8368196}.
}

{\color{black} {\bf Cooperation with Different Types of Vehicles.}} Recent work\footnote{As an example, see the DJI-Ford Challenge (http://developer.dji.com/challenge2016/) and the recent Ford's patent on self-driving vehicles with detachable drone (https://www.google.ch/patents/US9555885)} has explored the use of drones coupled with vehicles, e.g., where vehicles are drone stations, from which drones to survey disaster-struck areas are deployed. Drones can be used to guide self-driving vehicles or do delivery on behalf of such vehicles, or a collection of vehicles can band together to send a drone out to look ahead to observe in detail the  situation far ahead, or around the corner. {\color{black} Autonomous vehicles in the future might come with their own drones.\footnote{https://www.fastcompany.com/90162582/why-your-autonomous-car-might-come-with-its-own-drone}}

There are also other possibilities such as wheelchairs being integrated with  self-driving vehicles, so that an integrated system of door-to-door transportation can be made - e.g., an automated wheelchair coordinating with an automated  vehicle on where to drop-off and pick-up, to provide a complete mobility solution to the disabled, even ride-sharing solutions but for wheelchaired people.
Self-driving vehicles can also coordinate with not only people, but robots and things, for their pick-up or drop-off. A thing or robot might require repairs and so crowdsources  self-driving vehicles to send it to a repairer - all automated, if the owner pre-authorizes. An automated vehicle may be sent to pick up a broken down automated vehicle.
automated vehicles can also coordinate with fueling (or charging) stations automatically, in determining appropriate schedules in order to reduce wait times.  {\color{black} Automated flying taxis\footnote{See https://www.scientificamerican.com/article/heres-whats-needed-for-self-flying-taxis-and-delivery-drones-to-really-take-off/} might also cooperate with automated vehicles to deliver an end-to-end automated transport solution.}

{\color{black}
{\bf Summary.} Table~1 shows cooperative behaviours for vehicles and their approaches - achieving a particular behaviour might  involve not just hardware control actions and network communications,  but high-level cooperative protocols and reasoning at the software level. In particular, one can imagine a social brain module determining how the vehicle is to interact and respond to other vehicles. High-level reasoning about cooperative behaviour could translate into low-level vehicular  control actions (e.g., speed adjustments, gear changes, steering, manoeuvring, etc), say in parking, routing, dynamic traffic flow movements, platooning, and intersection passing, but cooperative awareness and social memory (and vehicle social networks) might not immediately translate into control actions, but into knowledge assimilation and formation of long term memories, respectively, which might only translate into control actions later. 

There are also cooperative behaviours which might not require high-level reasoning, e.g., certain forms of platooning (though it is noted that v2v communications, in addition to short-range sensing, has advantages for platooning~\cite{BERGENHEM20121222}) and certain swarm movements, where short-range sensing (e.g., inter-vehicle distances) and reactive control action behaviour, without the need for sophisticated high-level reasoning or planning, are adequate for a group of vehicles to behave in a certain cooperative manner.

}

\begin{table*}[ht] 
{\color{black}
\centering
\begin{tabular}{| p{3cm} | p{8cm} |  p{5cm} |}   \hline
{\bf Cooperative Behaviour}  &  {\bf Approaches  and Communications}  &  {\bf Selected References and Examples} \\ \hline \hline
Parking   & centralised and decentralised via v2v communications alone, or involving v2i communications with infrastructure help    & \cite{Kokolaki:2012:OAP:2161001.2161219,7457417,Bessghaier2012,7580732}     \\ \hline
Routing  &   centralised knowledge for coordination, v2v information sharing and routing, or indirect coordination methods such as stigmergy      &   \cite{5730496,DBLP:journals/tits/DesaiLDS13,10.1371/journal.pone.0159110,7313112}  \\ \hline
Dynamic Traffic Flows    &  swarm models of behaviour via v2v communications and VANETs &        \cite{riener,7553030,7328361,10.1007/978-3-319-72823-0_58}  \\ \hline
Platooning, Intersection and Safety   &  vehicle platooning protocols, use of intersection coordinators, or purely v2v communications to share status information for coordination, computation of individual vehicle traffic flows to avoid collisions, decentralised negotiation on road resources, cooperative adaptive cruise control methods  &         \cite{7547317,BERGENHEM20121222,DBLP:journals/comsur/JiaLWZS16,7934129,6629471,8023664,riener,7562449,4012536,4357693,7929602,5342530,8317924,6156445,7070682,6907953,6338827,6824011,ZHU2015363,6121907,LEE2013193,DBLP:journals/automatica/MalikopoulosCZ18,DBLP:journals/corr/abs-1708-06374,8370676,DBLP:conf/itsc/ManzingerA18} \\ \hline
Cooperative Awareness &  intelligent aggregation of information shared via v2v communications and high-level reasoning &  \cite{10.1007/978-3-642-41054-3_3,8119570,10.4108/eai.31-8-2017.153052} \\ \hline
Social Memory and Social Networks &  social networks among people in vehicles induced by proximity of vehicles, or social networks for vehicles capturing social relationships among vehicles (separate from human social networks) & \cite{Maglaras_2016,doi:10.1108/LHT-12-2017-0259,luan2015feel,6849025,BUTT201868,8368196} \\ \hline
Cooperation with Different Types of Vehicles &  the types of networking involved can be varied and both technology and use-cases are still open areas of research & with drones (e.g., Ford vehicles), wheelchairs and robots \\ \hline
\end{tabular}
\vspace{0.5cm}
\caption{Summary of Cooperative Behaviours for Vehicles and Relevant Work.}
}
\end{table*}

\subsection{Vehicle-to-Pedestrian Cooperation}

CAVs will share roads with pedestrians, so that the interaction between a vehicle and the humans outside the vehicle is an issue to address~\cite{urban,7501845,2018arXiv180511773R,8000288}.\footnote{See also http://urban-online.org/en/human-factors-in-traffic/index.html}   Human-vehicle interaction  mediated by v2x communication (e.g., vehicle to wearable device) can provide a greater level of communication between human and vehicle, complementary to, or as an alternative to, visual physical gestures or signaling. A Vehicle-to-Pedestrian framework is provided in~\cite{8082781} using a DSRC-enabled smartphone.

There is also opportunity for cars to cooperate in making provisions for humans, e.g., a collection of cars slow down to enable an elderly person to cross the road. Also, cars at a scene with lots of people can exchange information with one another along a road so that cars still far from the scene can ``see'' pedestrians that their own cameras cannot see. 
Such human-to-vehicle messaging can complement vision-based techniques for recognising pedestrian activities~\cite{8241847,4633644}.

\section{Challenges}

Below, we highlight a range of challenges for cooperative vehicles. We do not aim to be exhaustive but to be comprehensive.





 
 
 
 

 
 \subsection{Scales of Cooperation}
 
 Cooperation among vehicles has been formulated as a constrained optimal control problem, where a
performance criterion is optimized given the vehicles'  trajectories, and subject to safety
and liveness requirements~\cite{7736181}. However, on a larger scale with open systems of large numbers of vehicles, exact knowledge of requirements might be difficult to obtain in real-time.
A different formulation of the problem with respect to open systems of vehicles, where decentralized mechanisms with global (or large-scale) behaviour emerging from local interactions among small groups of vehicles
might be needed.

It is unlikely or unnecessary that all automated vehicles in a country or a city cooperate. Often, automated vehicles close to each other (e.g., a platoon) or within a locality (near an intersection, a car park or a stretch of a  road) might cooperate. It can be shown that such local cooperation can lead to global effects - for example, cars coordinating routes when they meet at intersections is adequate to reduce traffic  congestion on a much larger scale~\cite{DBLP:journals/tits/DesaiLDS13}.
Also, a set of cars cooperating to park more efficiently in a part of the Central Business District (CBD) could reduce traffic congestion for cars in other parts of the CBD. While there could be beneficial emergent effects on much larger scales, how one could engineer wide-scale benefits from disparate sets of vehicles cooperating locally is still an open research question. {\color{black}
An interesting question raised in~\cite{7562449} is what the minimum number of connected vehicles   
  need to be  in order ``to start realizing the potential benefits''.

A related issue is how fully connected automated vehicles will co-exist with other non-equipped vehicles. This could be a likely scenario since it requires time for already on-road vehicles to be modernised and replaced. Dedicated areas and lanes for fully automated vehicles separate from those for non-automated vehicles, as well as clear markings on automated vehicles, might be needed. Also, even among automated and driven vehicles, not all might participate in cooperative behaviours - human drivers can ignore cooperative behaviour recommendations, some vehicles might not be equipped with that ability, and  even automated vehicles might fail to cooperate. The work in~\cite{10.1371/journal.pone.0182621} and in~\cite{ALIEDANI2018} showed that even with not all the vehicles cooperating (e.g., 40\% to 70\%), travel time gains can still be obtained for routing, and time-to-park can still be reduced. Such work suggests that even with some vehicles cooperating, and not all, vehicles that cooperate can still benefit, though not as much as all vehicles cooperating, i.e. some cooperation is better than none. Also, it is not clear how much disadvantaged non-cooperative vehicles will be - while this might become an incentive to cooperate, not all vehicles might be equipped to do so for a given period. In a mixed environment with human-driven and fully automated vehicles, and not all vehicles being able to cooperate, the benefits of cooperation and the costs of non-cooperation (intentionally or not) need to be considered.
}






\subsection{Trusted Communications and Deception-Proofing}
It is also not clear when and which messages should be trusted. Vehicles can receive incident messages which are false and react in ways that endanger other vehicles. Vehicles need to take into account their own situation/context when responding to other messages. There has been recent work~\cite{6698224} on attempting to detect and deal with false or deceptive messages and verifying and validating messages in v2v communications, though not in the context of cooperation. Cooperative mechanisms should be relatively  robust and resilient to invalid messages or deception, or even the misinterpretation of particular vehicles of a received message.

As pointed out in~\cite{UllmannStrubbeWieschebrink2016}, while a CAM message provides a lot of information about a vehicle, such as 
 position, speed,  and direction at a specific
time, with precision in tens of centimeters, the receiver might not be able to trust the CAM data due
to  sensor inaccuracies, modifications of in-vehicle
components, and possible jamming attacks or transmission collisions which can delay messages.
Also, even if messages can be validated and  only truthful communications happen, there is a chance that some vehicles could behave in ways that are non-cooperative.
 Improvements are needed regarding the reliability and performance of the v2x communications to avoid or reduce lag time during cooperative manoeuvres, and there is a need for high accuracy of the data exchanged as well as high performance reasoning capabilities in vehicles. 
Also, vehicles can form coalitions that can preclude non-coalition vehicles from taking advantage of cooperation. Hence, cooperation will also need to be robust against malicious coalition behaviours. 
The work in~\cite{Chen2017} which showed that the  probability
of correct message delivery reduces to its minimum
after the proportion of malicious vehicles in
the network crosses a threshold, that is, trust requires a collective effort.

{\color{black}
Related to the issue of trust are also ethical issues concerning how vehicles might deal with false or misleading information. For example, in VANETs, would it be ethical for a vehicle to forward a message to another vehicle knowing that the message is false? Or what if a vehicle forwards a fake message but it does not know that it is fake? A question is whether each vehicle is required to use its time and resources  to verify or validate the veracity of a message before forwarding it - e.g., if a claimed message is from a police or emergency vehicle, while authentication is certainly sensible before acting on the message, a question is whether authentication should be required before it is forwarded to other vehicles.

Also, vehicles can receive information from different infrastructure (RSUs) at different locations - such information might need to be authenticated before being acted on or before being forwarded to  other vehicles. There could be different security policies for different zones the vehicles are in - e.g., a vehicle enters a private parking building, and is issued information which must then be authenticated, and when the same vehicle goes into a different car park, or zone, different policies might also apply. 
}

\subsection{Standards: Cooperation Protocols and Behaviour}
There have been efforts underway to harmonise Europe's CAM and the US SAE standards  for message data elements.\footnote{E.g., see the report by the Center for Automotive Reearch at http://www.cargroup.org/\-wp-content/uploads/\-2017/02/GLOBAL-HARMONIZATION-\-OF-CONNECTED-VEHICLE-COMMUNICATION-STANDARDS.pdf and https://docbox.etsi.org/workshop/\-2014/201402\_ITSWORKSHOP/\-S02\_ITS\_SomeBitsFromtheWorld/HONDA\_BAI.pdf}  {\color{black} The cost of equipping a vehicle with v2v communication capability must be offset by benefits, which are mainly experienced with more vehicles with such capability, so that vehicles by different manufacturers should be able to interact.}

There may also be a need for standards to define the language of messages that the  CAVs can understand as well as the operational meaning of the messages. The operational meaning  refers to how CAVs are to behave in response to such messages - a question is whether  one should standardise how CAVs respond, and not only the messages. For example, when receiving an alert about emergency messages, the behaviour of the CAVs must be to carefully give way.
There are questions about how the vehicles should cooperate if they are from different manufacturers -
for example, on  a message broadcast from an emergency vehicle, each vehicle seeks to get out of the way safely in a certain pre-programmed manner, but 
 a question is how to program each vehicle's behaviour to ensure that a clear way will emerge for the emergency vehicle to pass through.

There are also issues with human-vehicle interaction - what protocols are needed for human drivers to interact with automated vehicles, and for pedestrians to interact with automated vehicles, in a standard way (for vehicles from all manufacturers). By having clear expectation of behaviours of CAVs in different road situations, humans might then be better able to adapt their behaviour around CAVs. 

There could be a range of cooperation protocols for CAVs in different situations, e.g., a platooning protocol, a protocol for interaction at intersections, a protocol for directional swarming, a protocol for interaction in a car park, a protocol for traffic merging~\cite{Aoki:2017:MPS:3055004.3055028} and so on.  High-level representations of CAV cooperative behaviour using multiagent models could be explored and protocols formally verified, e.g., as done for platooning protocols using a Belief-Desire-Intention agent model of vehicles~\cite{KAMALI2017}, and using the pi-calculus to model v2v  communication protocols and low-level complex vehicle dynamics~\cite{7743450}. While it is convenient to study each protocol separately, there is an eventual need for each CAV to be able to cooperate in multiple situations, with multiple cooperation protocols  integrated into a  {\em social brain} module. {\color{black} Context-awareness, which takes into the account the vehicles current goals and current knowledge about the vehicle's situation, can be explored as a mechanism to trigger the selection and prioritization of the right cooperation protocol under different circumstances, or even the switching of cooperation protocols corresponding to the need to switch cooperative behaviours when  cooperating with different groups of vehicles in different road situations (e.g., from platooning on a highway to passing an intersection, and then merging into another lane and then cooperatively dropping off passengers).  }

{\color{black}
CAVs have the advantage of connectivity as opposed to AVs relying only on sensors of surroundings; there is considerable advantage of obtaining information from not only sensors but also from other vehicles and road-side-units via messages. However, as noted in the scenarios earlier, the advantage of connectivity is not just to receive information about the surroundings, but also in cooperating to achieve goals, or coordinating behaviours, e.g., in coordinating routes and in resolving contention for parking spots. Such cooperative behaviours will need to be implemented as possible actions in the social brain module of a vehicle.
}

\subsection{How Should Vehicles Talk to Each Other and with the Infrastructure?}

While standardized message formats (as we noted above) will help to enable interoperability, the way in which vehicles should or must respond to a given message remains difficult to specify. An interesting direction is to consider how  multiagent communication languages~\cite{Labrou:1999:ACL:631308.631316} developed for specifying meaningful interaction between software agents, based on speech act theory\footnote{https://plato.stanford.edu/entries/speech-acts/} can be reinterpreted in the context of v2x cooperation.

{\color{black}
It is also interesting to note above that many useful coordination schemes might use road-side-units (e.g., at intersections) while other schemes are totally decentralised, requiring only communications with peer vehicles. A question is whether both such schemes can coexist within a given area, and so, the vehicles will need to be able to talk different types of protocols, e.g., when at different intersections, or whether there will be standardised schemes.
}

\subsection{Context-Aware Decision-Making and Regulations}

The ability to understand the situation in particular regions is needed by CAVs  in decision making. Consider the following example first introduced in~\cite{DBLP:series/sbcs/Loke17}, but simplified here. 
A CAV can be programmed with a particular destination, and it could bring the passengers there, but upon arrival, 
 the car intends to drop off the passengers (including the driver) and then either proceeds to find a car park nearby, or  cruises around nearby.
 This is what we would expect but suppose the place turns out to have a traffic jam so that even simply dropping off the passengers would not be easy and the car might be stuck in traffic waiting for its turn at the drop-off zone.  
 So, the car could try to drop the passengers off a bit further away from the main drop-off zone, with approval from the driver~\cite{8370676}. 
Hence,  the car has to know when it needs to involve its passengers in such decision-making, even if it is assumed that  the passengers are simply leaving it to the car to take them to the right place. 

{\color{black}
Recent work explores the use of ontology-supported knowledge-based reasoning for vehicles to reason with data from sensors, maps, and driving paths, such as in~\cite{DBLP:journals/ieicet/ZhaoILMS17}, where such reasoning have been shown to enable autonomous vehicles to understand road situations
and make   decisions to avoid collisions.  Indeed, information from CAM messages and other map and vehicle information can be integrated in one place to enable a vehicle to reason more comprehensively about  its situation.
The concept of the Local Dynamic Map (LDM)\footnote{https://www.etsi.org/deliver/etsi\_tr/102800\_102899/102863/\-01.01.01\_60/tr\_102863v010101p.pdf} acts as a platform to combine
static   geographic information system (GIS) maps, with  objects (e.g., vehicles and pedestrians) which are dynamic in nature. The work in~\cite{Eiter2018} provided an LDM ontology allowing knowledge-based integration and querying of LDM knowledge to support situation awareness, e.g., to query about surrounding vehicles in safety applications. 
 }

\subsection{Lawful Interactions}

Often, the vehicles will need to follow traffic regulations and laws concerning what to do on receiving incident messages. For example, a recent road rule in Melbourne, Australia, is that vehicles must slow to no more than 40km/h when passing crashes and incidents where emergency vehicles are stopped on the side of the road. Hence, CAVs will need to be able to respond to that. In many countries,  vehicles must make way for a government vehicle or an emergency vehicle passing through. It is not clear CAVs need to be programmed to be context/situation-aware and still respond safely to receiving messages. There has been recent attention on robot laws and building ethical or lawful behaviour into AI~\cite{roboethics}, though not focused on CAVs.
 
Situational rules might need to be represented within such vehicles, e.g., the fact that certain vehicles have certain rights (such as right of way) or certain obligations in particular contexts~\cite{searle2010making}:
$X$ (in certain physical world positions) {\bf counts as} $Y$ (special car status with certain rights) {\bf in} $C$ (some real world conditions).

There has been work on norms and electronic institutions from multiagent research (e.g.,~\cite{Noriega2016}), with interesting  applications to creating e-institutions for CAVs in order to regulate CAV behaviours.

\section{Summary and Conclusion}

CAVs can cooperate to improve safety and efficiency in a wide variety of situations, but this calls for an AI of  cooperation in vehicles. 
This paper has outlined the enormous  possibilities for transforming transport with cooperative CAVs, with a focus on cooperative social intelligence and reasoning needed for vehicles.

It is not so clear which cooperative behaviours should be best implemented at the reactive control level only and which requires higher-level situation-aware reasoning.
Brooks' ~\cite{1087032} layered subsumption architecture and newer conceptualizations of layered robot architectures such as \cite{DBLP:journals/corr/abs-1811-03563} may offer a direction of thinking about a layered architecture for combining cooperative behaviours (within a social brain for connected automated vehicles), thereby  providing suitable social-AI behaviours for a wide range of scenarios.
Complexity science and swarm computing, where individual vehicles follow particular rules and interact locally, yielding wide-ranging emergent behaviours can also be explored for scalable  coordination of vehicles.
There are also   issues that need to be addressed via inter-disciplinary  research for the benefits to be fully realized.  The {\em social brain} of the socially intelligent vehicle will complement the close proximity sensing and  intelligence that is already being built into CAVs today. Automated vehicles is only one of many large scale possibilities with the Internet of autonomous things in public - one can imagine cities saturated with robotic things which will need to cooperate and face many similar issues.

{\color{black}
\section*{Acknowledgements}
The author would like to thank the anonymous reviewers for their valuable comments which helped  improve this paper.
}

\bibliographystyle{unsrt}
\bibliography{magazine}

\vspace{-1cm} 
\begin{IEEEbiography}[{\includegraphics[width=1in,height=1.25in,clip,keepaspectratio]{sengloke}}] {Dr. Seng W. Loke}
  is Professor in Computer Science at the School of Information Technology in Deakin University. He received his Ph.D. degree  in Computer Science from the University of Melbourne, Australia, in 1998.  He has authored and co-authored over 300 publications. He also authored ``Context-Aware Pervasive Systems: Architectures for a New Breed of Applications''  by Auerbach (CRC Press), Dec 2006, and   ``Crowd-Powered Mobile Computing and Smart Things'' published by Springer, 2017.  
His research has mainly been in Internet of Things (IoT), pervasive   and mobile computing,  intelligent transport systems, with current focus on complex trusted cooperation among Things (including smart cooperative vehicles),  and  the social impact of   technology innovation.
\end{IEEEbiography}

\end{document}